\documentclass[final]{cvpr}

\usepackage{times}
\usepackage{epsfig}
\usepackage{graphicx}
\usepackage{amsmath}
\usepackage{amssymb}
\usepackage{subcaption}
\usepackage{booktabs}
\usepackage{tabularx}

\usepackage{color} 
\usepackage{setspace}
\usepackage[frozencache=true,cachedir=.]{minted}

\makeatletter
\@namedef{ver@everyshi.sty}{}
\makeatother
\usepackage{tcolorbox}
\usepackage{amsthm}

\theoremstyle{definition}
\newtheorem*{definition}{Definition}

\tcbuselibrary{listings, minted, skins}

\tcbset{listing engine=minted}

\definecolor{mygray}{rgb}{0.99, 0.99, 0.99}
\setminted[python]{fontsize=\footnotesize, baselinestretch=1.0, linenos, xleftmargin=1em}
\setminted[Text]{fontsize=\footnotesize, baselinestretch=1.0,
linenos, xleftmargin=1em}

\usepackage[pagebackref=true,breaklinks=true,colorlinks,bookmarks=false]{hyperref}

\def\cvprPaperID{37} 
\def\confYear{CVPR 2021}

\usepackage{xcolor}
\usepackage{enumitem}
\usepackage{inconsolata}

\newcommand{\codeinline}[1]{\texttt{#1}}

\begin{document}

\title{Avalanche: an End-to-End Library for Continual Learning}


\author{Vincenzo Lomonaco$^{1 \dagger}$\thanks{Corresponding author: vincenzo.lomonaco@unipi.it}~
Lorenzo Pellegrini$^{2}$\thanks{Avalanche lead maintainers.}
Andrea Cossu$^{1}$$^,$$^{18}$\footnotemark[2]~
Antonio Carta$^1$\footnotemark[2]~
Gabriele Graffieti$^{2}$\footnotemark[2]\\
Tyler L. Hayes$^{3}$~
Matthias De Lange$^{4}$~
Marc Masana$^{5~}$~
Jary Pomponi$^{6~}$~
Gido van de Ven$^{7~}$\\
Martin Mundt$^{8~}$~
Qi She$^{9~}$~
Keiland Cooper$^{10~}$~
Jeremy Forest$^{11~}$~
Eden Belouadah$^{12~}$\\
Simone Calderara$^{13~}$~
German I. Parisi$^{14~}$~
Fabio Cuzzolin$^{15~}$~
Andreas Tolias$^{7~}$~
Simone Scardapane$^{6~}$\\
Luca Antiga$^{16~}$~
Subutai Amhad$^{17~}$~
Adrian Popescu$^{12~}$~
Christopher Kanan$^{3~}$\\
Joost van de Weijer$^{5~}$~
Tinne Tuytelaars$^{4~}$~
Davide Bacciu$^{1~}$~
Davide Maltoni$^{2~}$\vspace{0.1cm}\\
  {\normalsize
  $^1$University of Pisa \quad
  $^2$University of Bologna \quad
  $^3$Rochester Institute of Technology}\\
  {\normalsize
  $^4$KU Leuven \quad
  $^5$Universitat Autònoma de Barcelona \quad
  $^6$Sapienza University of Rome}\\
  {\normalsize
  $^7$Baylor College of Medicine \quad
  $^8$Goethe University \quad
  $^9$ByteDance AI Lab}\\
  {\normalsize
  $^{10}$University of California \quad
  $^{11}$New York University \quad
  $^{12}$Université Paris-Saclay}\\
  {\normalsize
  $^{13}$University of Modena and Reggio-Emilia \quad
  $^{14}$University of Hamburg}\\
  {\normalsize
  $^{15}$Oxford Brookes University\quad
  $^{16}$Orobix\quad
  $^{17}$Numenta\quad
  $^{18}$Scuola Normale Superiore\vspace{0.1cm}}\\
  {\url{https://avalanche.continualai.org}}
}

\maketitle

\begin{abstract}
    
Learning continually from non-stationary data streams is a long-standing goal and a challenging problem in machine learning. Recently, we have witnessed a renewed and fast-growing interest in continual learning, especially within the deep learning community. However, algorithmic solutions are often difficult to re-implement, evaluate and port across different settings, where even results on standard benchmarks are hard to reproduce. In this work, we propose Avalanche, an open-source end-to-end library for continual learning research based on PyTorch. Avalanche is designed to provide a shared and collaborative codebase for fast prototyping, training, and reproducible evaluation of continual learning algorithms.
    
\end{abstract}

\section{Introduction}
\emph{Continual Learning} (CL), also referred to as \emph{Lifelong} or \emph{Incremental Learning}, is a challenging research problem~\cite{chen2018}. Lately, it has become the object of fast-growing interest from the research community, especially thanks to recent investigations leveraging gradient-based deep architectures~\cite{hadsell2020, parisi2019}. In the last few years, machine learning has witnessed a prolific, variegated, and original research production on the topic: from Computer Vision~\cite{delange2021continual, lomonaco2019, masana2020class} to Robotics~\cite{lesort2020, pellegrini2019latent, parisi2020online}, from Reinforcement Learning ~\cite{khetarpal2020towards, lomonaco2020} to Sequence Learning~\cite{cossu2021}, among others.

\begin{figure}[t]
\begin{center}
  \includegraphics[width=\columnwidth]{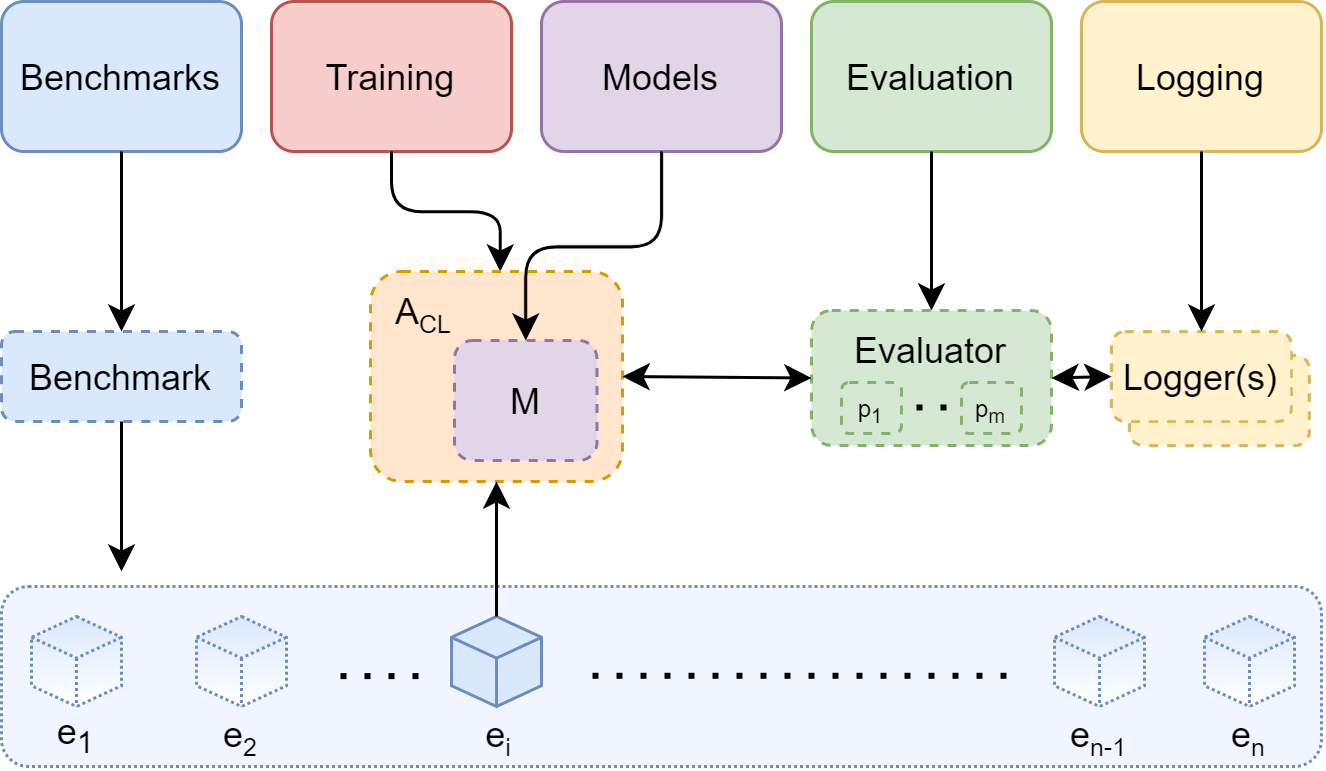}
\end{center}
  \caption{Operational representation of Avalanche with its main modules (top), the main object instances (middle) and the generated stream of data (bottom). A \emph{Benchmark} generates a stream of experiences $e_i$ which are sequentially accessible by the continual learning algorithm $A_{CL}$ with its internal model $M$. The \emph{Evaluator} object directly interacting with the algorithm provides a unified interface to control and compute several performance metrics ($p_i$), delegating results logging to the \emph{Logger(s)} objects.}
\label{fig:avl}
\end{figure}

However, continual learning algorithms today are often designed and implemented from scratch with different assumptions, settings, and benchmarks that make them difficult to compare among each other or even port to slightly different contexts. A crucial factor for the consolidation of a fast-growing research topic in the machine learning domain is the availability of tools and libraries easing the implementation, assessment, and replication of models across different settings, while promoting the reproducibility of results from the literature~\cite{pineau2020improving}.

In this work, we propose \emph{Avalanche}, an open-source (MIT licensed) end-to-end library for continual learning based on PyTorch~\cite{paszke2019pytorch}, devised to provide a shared and collaborative codebase for fast prototyping, training, and evaluation of continual learning algorithms.

We designed Avalanche according to a set of fundamental design principles (Sec.~\ref{sec:principles}) which, we believe, can help researchers and practitioners in a number of ways: i) \emph{Write less code, prototype faster, and reduce errors}; ii) \emph{Improve reproducibility}; iii) \emph{Favor modularity and reusability}; iv) \emph{Increase code efficiency, scalability and portability}; v) \emph{
Foster impact and usability}.

The contributions of this paper can be summarized as follows:

\begin{enumerate}[noitemsep, nolistsep]
    \item We propose a general continual learning framework that provides the conceptual foundation for Avalanche (Sec.~\ref{sec:framework}).
    \item We discuss the general design of the library based on five main modules: \emph{Benchmarks}, \emph{Training}, \emph{Evaluation}, \emph{Models}, and \emph{Logging} (Sec.~\ref{sec:modules}).
    \item We release the open-source, collaboratively maintained project at \url{https://github.com/ContinualAI/avalanche}, as the result of a collaboration involving over 15 organizations across Europe, United States, and China.
\end{enumerate}

Prior work related to this project is discussed in Section~\ref{sec:related}. Lastly, we discuss in Section~\ref{sec:conclusion} the importance that an \emph{all-in-one}, \emph{community-driven} library may have for the future of continual learning and its possible extensions.

\section{Design Principles}
\label{sec:principles}

Avalanche has been designed with five main principles in mind: i) \emph{Comprehensiveness and Consistency}; ii) \emph{Ease-of-Use}; iii) \emph{Reproducibility and Portability}; iv) \emph{Modularity and Independence}; v) \emph{Efficiency and Scalability}. These principals, we argue, are important for any continual learning project but become essential for tackling the most interesting research challenges and real-world applications.

\paragraph{Comprehensiveness and consistency} The main design principle for Avalanche follows from the concept of \emph{comprehensiveness}, the idea of providing 
an exhaustive
and unifying library with \emph{end-to-end} support for continual learning research and development. A comprehensive codebase does not only provide a unique and clear access point to researchers and practitioners working on the topic, but also favors \emph{consistency} across the library,  with a coherent and easy interaction across modules and sub-modules. Last but not least, it promotes the consolidation of a large community able to provide better support for the library.

\paragraph{Ease-of-Use} The second principle is the focus on \emph{simplicity}: simple solutions to complex problems and a simple usage of the library. We concentrate our efforts on the design of an intuitive Application Programming Interface (API), an official website, and rich documentation with a curated list of executable notebooks and examples\footnote{The official website, documentation, notebooks, and examples are available at \url{https://avalanche.continualai.org}.}.

\paragraph{Reproducibility and Portability} Reproducing research paper results is a difficult but much needed task in machine learning~\cite{hutson2018artificial}. The problem is exacerbated in continual learning. A critical design objective of Avalanche is to allow experimental results to be seamlessly reproduced. This allows researchers to simply integrate their own original research into the shared codebase and compare their solution with the existing literature, forming a virtuous circle. Hence, reproducibility is not only a core objective of sound and consistent scientific research, but also a means to speed up the development of original continual learning solutions.

\paragraph{Modularity and Independence} \emph{Modularity} is another fundamental design principle. In Avalanche, simplicity is sometimes bent in favor of modularity and reusability. This is essential for scalability and to collaboratively bring the codebase to maturity. A particular focus on module \emph{independence} is maintained to guarantee the stand-alone usability of individual module functionalities and facilitates learning of a particular tool.

\paragraph{Efficiency and Scalability} Computational and memory requirements in machine learning have grown significantly throughout the last two decades~\cite{thompson2020computational}. Standard deep learning libraries such as TensorFlow~\cite{tensorflow2015-whitepaper} or PyTorch~\cite{paszke2019pytorch} already focus on \emph{efficiency} and \emph{scalability} as two fundamental designing principles, since modern research experiments can take months to complete~\cite{schwartz2019green}. Avalanche is based on the same principles: offering the end-user a seamless and transparent experience regardless of the use-case or the hardware platform that the library is run on.

\section{Continual Learning Framework}
\label{sec:framework}

Recently, we have witnessed a significant attempt to formalize a general framework for continual learning algorithms~\cite{lomonaco2019, lomonaco2020, vandeven2018a}. These proposals often categorize scenarios and algorithms based on their unique properties and specific settings. However, as outlined in this paper, within the formal design of Avalanche, we take a different approach. 

Given the fast-evolving, often conflicting views of the problem, we aim to lower the number of assumptions to a minimum, favoring simplicity and flexibility. In practice, this translates into providing users with a set of building blocks that can be used in any continual learning solution without imposing any strong nomenclature, constraining abstractions, or assumptions.

In Avalanche, data is modeled as an ordered sequence (or stream) composed of $n$, usually \emph{non-iid}, learning \emph{experiences}: 

\begin{equation*}
    e_0, e_1, \dots, e_n.
\end{equation*}
A learning experience is a set composed of one or multiple samples which can be used to update the model. This is often referred to in the literature as \emph{batch} or \emph{task}. 
This formulation is general enough to be used in several continual learning contexts, such as supervised, reinforcement, or unsupervised continual learning. Avalanche provides a general set of abstractions that do not impose any particular constraints on the content of the experiences. For example, in a supervised training regime, each learning experience $e_i$ can be seen as a set of triplets $\langle x_i, y_i, t_i \rangle$, where $x_i$ and $y_i$ represent an input and its corresponding target, respectively, while $t_i$ is the \emph{task label}, which may or may not be available. 

During training, a \emph{continual learning algorithm} $A_{CL}$ processes experiences sequentially and uses them to update the model and its internal state. In Avalanche, each algorithm has a training mode, used to update the model, and an evaluation mode, which may be used to process streams of experiences for testing purposes.

The continual learning framework we propose can be formalized as follows. 

\begin{definition}[Continual Learning Framework]
A continual Learning algorithm $A_{CL}$ is expected to update its internal state (e.g. its internal model $M$ or other data structures) based on the sequential
exposure to a non-stationary stream of experiences ($e_1, \dots, e_n$). The objective of $A_{CL}$ is to improve its performance on a set of metrics ($p_1,\dots,p_m$) as assessed on a test stream of experiences ($e_1^{t}, \dots, e_n^t$). 
\end{definition}

\section{Main Modules}
\label{sec:modules}

The library is organized into five main modules: \textbf{Benchmarks} (Sec. \ref{sec:benchmarks}), \textbf{Training} (Sec. \ref{sec:training}), \textbf{Evaluation} (Sec. \ref{sec:eval}), \textbf{Models} (Sec. \ref{sec:models}), and \textbf{Logging} (Sec. \ref{sec:logging}). 
Table~\ref{tab:features} summarizes the features provided by Avalanche at the current stage of development. In Fig.~\ref{fig:avl}, an operational representation of the library modules and their interplay within the aforementioned reference framework is shown.






\begin{table*}[t]
    \renewcommand{\arraystretch}{0.9}
    \centering
    \begin{tabularx}{\textwidth}{c|X}
    \toprule
    & \multicolumn{1}{c}{\textbf{Supported features}}  \\
    \midrule
    \textbf{Benchmarks} & Split/Permuted/Rotated MNIST~\cite{lopez-paz2017}, Split Fashion Mnist~\cite{farquhar2018}, Split Cifar10/100/110~\cite{rebuffi2017,maltoni2019}, Split CUB200, Split ImageNet~\cite{rebuffi2017}, Split TinyImageNet~\cite{delange2021continual}, Split/Permuted/Rotated Omniglot~\cite{schwarz2018}, CORe50~\cite{lomonaco2017}, OpenLORIS~\cite{she2019}, Stream51~\cite{roady2020stream}.\\
    \textbf{Scenarios} & Multi Task~\cite{lesort2020}, Single Incremental Task~\cite{lesort2020}, Multi Incremental Task~\cite{lesort2020}, Class incremental~\cite{rebuffi2017icarl,vandeven2018a}, Domain Incremental~\cite{vandeven2018a}, Task Incremental~\cite{vandeven2018a}, Task-agnostic, Online, New Instances, New Classes, New Instances and Classes.  \\
    \textbf{Strategies} & Naive (Finetuning), CWR*~\cite{lomonaco2020a}, Replay, GDumb~\cite{prabhu2020}, Cumulative, LwF~\cite{li2016}, GEM~\cite{lopez-paz2017}, A-GEM~\cite{chaudhry2019}, EWC~\cite{kirkpatrick2017}, Synaptic Intelligence~\cite{zenke2017}, AR1~\cite{maltoni2019}, Joint Training. \\
    \textbf{Metrics} & Accuracy, Loss (user specified), Confusion Matrix, Forgetting, CPU Usage, GPU usage, RAM usage, Disk Usage, Timing, Multiply, and Accumulate~\cite{jeangoudoux2018,diaz-rodriguez2018}. \\
    \textbf{Loggers} & Text Logger, Interactive Logger, Tensorboard Logger~\cite{tensorflow2015-whitepaper}, Weights and Biases (W\&B) Logger~\cite{wandb} (in progress).\\
   \bottomrule
    \end{tabularx}
    \caption{Avalanche supported features for the Alpha release (v0.0.1).}
    \label{tab:features}
\end{table*}

\begin{figure}[t]
  \includegraphics[width=\columnwidth]{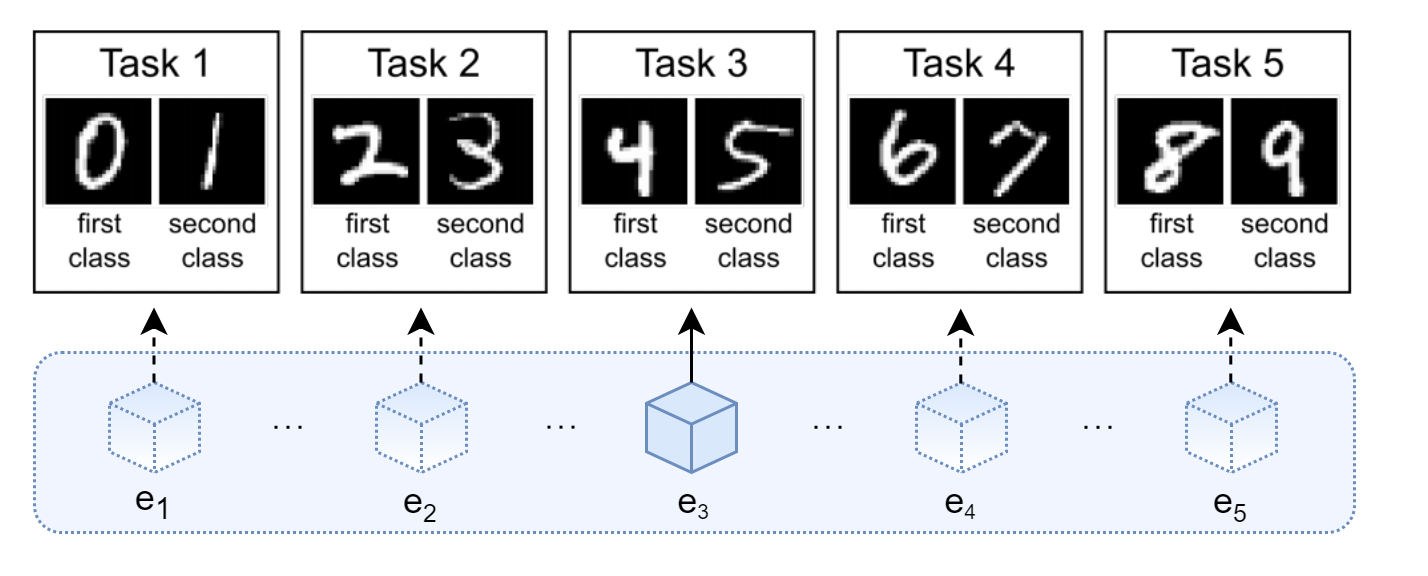}
  \caption{Example of a generated stream in Avalanche, composed by five experiences, implementing the common SplitMNIST benchmark \cite{zenke2017}. When accessing experience $e_3$, the $A_{CL}$ algorithm has no access to previous or future experiences.} 
\label{fig:exps_example}
\end{figure}

\subsection{Benchmarks}
\label{sec:benchmarks}

Continual learning revolves around the idea of dealing with a non-stationary stream of experiences. An example stream from the standard SplitMNIST benchmark \cite{zenke2017} composed of five experiences is shown in  Fig.~\ref{fig:exps_example}. A target system powered by a continual learning strategy is required to learn from experiences (e.g., by considering additional classes in a class-incremental setting \cite{masana2020class}) in order to improve its performance or expand its set of capabilities.
This means that the component in charge of generating the data stream is usually the first building block of a continual learning experiment. It is no surprise that a considerable amount of time is spent defining and implementing the data loading module.
The benchmarks module 
offers a powerful set of tools one can leverage to greatly simplify this process.

The term \emph{benchmark} is used in Avalanche to describe a recipe that specifies how the stream of data is created by defining the originating dataset(s), the contents of the stream, the amount of examples, task labels and boundaries~\cite{aljundi2019d}, etc. When defining such elements, some degree of freedom is retained to allow obtaining different \emph{benchmark instances}. For example, different instantiations of the SplitMNIST benchmark \cite{zenke2017} can be obtained by setting different class assignments. Alternatively, distinct instances of the PermutedMNIST~\cite{goodfellow2013empirical} benchmark can be obtained by choosing different pixel permutations.

The \emph{benchmarks} module is designed with the idea of providing an extensive set of out-of-the-box loaders covering the most common benchmarks (i.e. SplitCIFAR~\cite{rebuffi2017}, PermutedMNIST~\cite{goodfellow2013empirical}, etc.) through the \emph{classic} submodule. A simple example illustrating how to use the “SplitMNIST” benchmark \cite{zenke2017} is shown in Fig.~\ref{code:classic-benchmarks}.
Moreover, a wide range of tools are available that enables the creation of customized benchmarks. The goal is to provide full support to researchers implementing benchmarks that do not easily fit into the existing categories.

Most out-of-the-box benchmarks are based on the dataset implementation provided by the \codeinline{torchvision} library. A proper implementation is provided for other datasets (such as CORe50~\cite{lomonaco2017}, Stream-51~\cite{roady2020stream}, and OpenLORIS~\cite{she2019}). The benchmark preparation and data loading process can seamlessly handle memory-intensive benchmarks, such as Split-ImageNet~\cite{rebuffi2017}, without the need to load the whole dataset into memory in advance.

Further, the benchmarks module is entirely standalone, meaning that it can be used independently from the rest of Avalanche.

\paragraph{Benchmarks creation}

The benchmarks module exposes a uniform API that makes it easy to define a new continual learning benchmark.

\begin{figure}
\begin{tcolorbox}[title={Classic Benchmarks}, colback=mygray]
\begin{minted}{python}
benchmark_instance = SplitMNIST(
    n_experiences=5, seed=1)
    # Other useful parameters
    # return_task_id=False/True
    # fixed_class_order=[5, 0, 9, ...]
\end{minted}
\end{tcolorbox}
\caption{Simple instantiation of a \emph{Classic} continual learning benchmark.}
\label{code:classic-benchmarks}
\end{figure}

The \emph{classic} package hosts an ever-growing set of common benchmarks and is expected to cover the usage requirements of the vast majority of researchers. However, there are situations in which implementing a novel benchmark is required. Avalanche offers a flexible API that can be used to easily handle this situation.

Starting from the higher-level API, Avalanche offers explicit support for creating benchmarks that fit one of the ready-to-use scenarios. The concept of \emph{scenario} is slightly different from that of `benchmark' as it describes a more general recipe independent of a specific dataset.
If the benchmark to be implemented fits either in the \textit{New Instances} or \textit{New Classes} scenarios \cite{maltoni2019}, one can consider using one of the specific generators \codeinline{nc\_scenario} or \codeinline{ni\_scenario}. Both generators take a pair of train and test datasets and produce a benchmark instance. The New Classes generator splits all the available classes in a number of subsets equal to the required number of experiences. Patterns are then allocated to each experience by assigning all patterns of the selected classes. This means that the New Classes generator can be used as a basis to set up Task or Class-incremental learning benchmarks~\cite{vandeven2018a}. The New Instances generator splits the original training set by creating experiences containing an equal amount of patterns using a random allocation schema. The main intended usage for this generator is to help in setting up Domain-Incremental learning benchmarks~\cite{vandeven2018a}. Most classic benchmarks are based on these generators. Fig.~\ref{code:generators} shows a simple example of using \codeinline{nc\_scenario}.

\begin{figure}[ht]
\begin{tcolorbox}[title={Benchmarks Generators}, colback=mygray]
\begin{minted}{python}
# Nearly all datasets from torchvision
# are supported

mnist_train = MNIST('./mnist', train=True)
mnist_test = MNIST('./mnist', train=False)
benchmark_instance = nc_scenario(
    train_dataset=mnist_train,
    test_dataset=mnist_test,
    n_experiences=n_experiences,
    task_labels=True/False)
\end{minted}
\end{tcolorbox}
\caption{Example of the \emph{"New Classes"} benchmark generator on the MNIST dataset.}
\label{code:generators}
\end{figure}

If the benchmark does not fit into a predefined scenario, a \textit{generic generator} can be used. At the moment, Avalanche allows users to create benchmark instances from lists of files, Caffe-style filelists~\cite{jia2014caffe}, lists of PyTorch datasets, or even directly from Tensors. We expect that the number of generic generators will rapidly grow in order to cover the most common use cases and allow for maximum flexibility.

\paragraph{Streams and Experiences}

Not all continual learning benchmarks limit themselves to describing 
a single stream of data. Many 
contemplate an out-of-distribution stream, a validation stream and possibly several other arbitrary streams, each linked to a different semantic. For instance, \cite{chaudhry2019} proposes a benchmark where a separate stream of data is used for cross-validation, while 
\cite{roady2020stream} defines an out-of-distribution stream used to evaluate the novelty detection capabilities of the model.

Motivated by this remark,
we decided to model benchmark instances as a composition of streams of experiences. This choice has two positive effects on the resulting API. Firstly, the way streams and experiences can be accessed is shared across all benchmark instances. Secondly, this modeling of benchmark instances does not force any preconceived schema upon researchers. Avalanche leaves the semantic aspects regarding the definition and usage of each stream to the creator of the benchmark.

A simple example showing the versatility of this design choice concerns the test stream: in order to allow for a proper evaluation of a continual learning strategy, benchmarks do not only need to describe the stream of training experiences but also need to 
properly describe a testing protocol. Such protocol is, in turn, based on one or more test datasets on which appropriate metrics can be computed. 
In many cases, the test data may need to be structured into a sequence of `test experiences', analogously to what happens with the training data stream.
For instance, in Class-Incremental learning 
the test set may be split into different experiences each containing only test patterns 
related to classes 
in the corresponding training experience.


Avalanche currently supports two different streams: \textit{train} and \textit{test}, while the support for arbitrary streams (for instance, out-of-distribution stream) will be implemented in the near future.

Each experience can be obtained by iterating over one of the available streams. Fig.~\ref{code:train-loop} shows how, starting from a benchmark instance, streams can be retrieved and used. Each experience carries a PyTorch dataset, task label(s) and other useful benchmark-specific information that can be used for introspection. An experience also carries a numerical identifier that defines its position in the originating stream. In fact, experiences in a stream can be also accessed by index. This functionality makes it easy to couple related experiences from different streams.

\begin{figure}[ht]
\begin{tcolorbox}[title={Main Training Loop}, colback=mygray]
\begin{minted}{python}
train_stream = benchmark_instance.train_stream
test_stream = benchmark_instance.test_stream

for idx, experience in enumerate(train_stream):
    dataset = experience.dataset
    
    print('Train dataset contains', 
        len(dataset), 'patterns')

    for x, y, t in dataset:
        ...

    test_experience = test_stream[idx]
    cumulative_test = test_stream[:idx+1]
\end{minted}
\end{tcolorbox}
\caption{Example of the main training loop over the stream of experiences.}
\label{code:train-loop}
\end{figure}

\paragraph{Task Labels and Nomenclature}

Every mechanism, internal aspect, name of function and class in the benchmarks module were designed with the intent of keeping Avalanche as neutral as possible with respect to the presence of task labels. Task boundaries, task descriptors and task labels are widely used in the continual learning literature to define both semantic and practical aspects of a benchmark. However, the meaning of those concepts is usually blurred with the definition of the specific benchmark to which they are applied to, making it hard to clearly pin-point a generic way to manage them.

Based on this observation, and due to the fact that the usage of task-specific information may become more extravagant or sophisticated in the future, we decided that Avalanche should not force any kind of convention. This means that the choice of whether to use task labels and how to use them is left to the user.

Following this idea, the \codeinline{GenericCLScenario} class, which is the common class for all scenarios instances, allows researchers to assign task labels at pattern granularity, thus allowing for experiences with zero or more task labels. We deemed this the most natural choice for Avalanche: we believe that a continual learning library should not constrain researchers by superimposing a certain view of the field upon them. 
Instead, the idea of enabling the user to create complex setups in a simple way, without forcing a subjective interpretation, will probably prove to be more robust as the field continues to evolve.

\subsection{Training}
\label{sec:training}

The \codeinline{training} module implements both popular continual learning strategies and a set of abstractions that make it easy for a user to implement custom continual learning algorithms. Each strategy in Avalanche implements a method for training (\codeinline{train}) and a method for evaluation (\codeinline{eval}), which can work either on single experiences or on entire slices 
of the data stream. Currently, Avalanche provides $11$ continual learning strategies (with many more to come), that can be used to train baselines in a few lines of code, as shown in Fig.~\ref{code:strategies}. See Table~\ref{tab:features} for a complete list of the available strategies.

\begin{figure}[ht]
\begin{tcolorbox}[title={Training Strategies}, colback=mygray]
\begin{minted}{python}
strategy = Replay(model, optimizer, 
                  criterion, mem_size)
for train_exp in scenario.train_stream:
    strategy.train(train_exp)
    strategy.eval(scenario.test_stream)
\end{minted} 
\end{tcolorbox}
\caption{Simple instantiation of an already available strategy in Avalanche.}
\label{code:strategies}
\end{figure}

\paragraph{Training/Eval Loops} In Avalanche, continual learning strategies subclass \codeinline{BaseStrategy}, which provides generic training and evaluation loops. 
These can be extended and adapted by each strategy. For example, \codeinline{JointTraining} implements offline training by concatenating the entire data stream in a single dataset and training only once. The pseudo-code in Fig.~\ref{code:train-structure} shows part of the \texttt{BaseStrategy.train} loop (\texttt{eval} has a similar structure).

\begin{figure}[ht]
\begin{tcolorbox}[title={Training Structure}, colback=mygray]
\begin{minted}{python}
def train(experiences):
	before_training()
	for exp in experiences:
		train_exp(exp)
	after_training()

def train_exp(exp):
	adapt_train_dataset()
	make_train_dataloader()
	before_training_exp()
	for epoch in range(n_epochs):
		before_training_epoch()
		training_epoch()
		after_training_epoch()
	after_training_exp()
\end{minted} 
\end{tcolorbox}
\caption{Main training structure, the skeleton of the \texttt{BaseStrategy} class.}
\label{code:train-structure}
\end{figure}

Under the hood, \codeinline{BaseStrategy} deals with most of the boilerplate code. 
The generic loops are able to seamlessly handle common continual learning scenarios, independently of differences such as the presence or absence of task labels.

\paragraph{Plugin System} \codeinline{BaseStrategy} provides a simple callback mechanism. This is used by strategies, metrics, and loggers to interact with the training loop and execute their code at the correct points using a simple interface and provides an easy interface to implement new strategies by adding custom code to the generic loops. \codeinline{BaseStrategy} provides the global state (current mini-batch, logits, loss, ...) to suitable plugins so that they can access all the information they need. In practice, most strategies in Avalanche are implemented via plugins. This approach has several advantages compared to a custom training loop. Firstly, the readability of the code is enhanced since most strategies only need to specify a few methods. Secondly, this allows for multiple strategies to be combined together. For example, we can define a hybrid strategy that uses Elastic Weight Consolidation (EWC)~\cite{kirkpatrick2017} and replay using the snippet of code shown in Fig.~\ref{code:hybrid-strategies}. 

\begin{figure}[ht]
\begin{tcolorbox}[title={Hybrid Strategies}, colback=mygray]
\begin{minted}{python}
replay = ReplayPlugin(mem_size)
ewc = EWCPlugin(ewc_lambda)
strategy = BaseStrategy(
    model, optimizer, 
    criterion, mem_size,
    plugins=[replay, ewc])
\end{minted}
\end{tcolorbox}
\caption{Example of an on-the-fly instantiation of hybrid strategies through Plugins.}
\label{code:hybrid-strategies}
\end{figure}

\subsection{Evaluation} \label{sec:eval}

The performance of a CL algorithm should be assessed by monitoring multiple aspects of the computation~\cite{lesort2020}. The \codeinline{evaluation} module offers a wide set of metrics to evaluate experiments.\\
Avalanche's design principle is to separate the issues of \textit{what to monitor} and \textit{how to monitor} it: the \codeinline{evaluation} module provides support for the former through the metrics, while the \codeinline{logging} module fulfills the latter (Section~\ref{sec:logging}). Metrics do not specify in which format their output should be displayed, while loggers do not alter metrics logic. Metrics can work even without a logger: the strategy's \emph{train} and \emph{eval} methods return a dictionary with all the metrics logged during the experiment. 

Few lines of code are sufficient to monitor a vast set of metrics: \textit{accuracy}, \textit{loss}, \textit{catastrophic forgetting}, \textit{confusion matrix}, \textit{timing}, \textit{ram/disk/CPU/GPU usage} and Multiply and Accumulate~\cite{jeangoudoux2018} (which measures the computational cost of the model's forward pass in terms of floating point operations).
Each metric comes with a standalone class and a set of plugin classes aimed at emitting metric values on specific moments during training and evaluation.

\paragraph{Stand-alone Metrics}
Stand-alone metrics are meant to be used independently of all Avalanche functionalities. Each metric can be instantiated as a simple Python object.  The metric will keep an internal state to store metric values. The state can be reset, updated or returned to the user by calling the related \codeinline{reset}, \codeinline{update} and \codeinline{result} methods, respectively.

\paragraph{Plugin Metrics}
Plugin metrics are meant to be easily integrated into the Avalanche training and evaluation loops. Plugin metrics emit a curve composed by multiple values at different points in time. Usually, plugin metrics emit values after each training iteration, training epoch, evaluation experience or at the end of the entire evaluation stream. For example, \codeinline{EpochAccuracy} reports the accuracy over all training epochs, while \codeinline{ExperienceLoss} produces as many curves as the number of experiences. Each curve monitors the evaluation accuracy of an experience at the end of each training loop. \\ 
Avalanche recommends the use of already implemented helper functions to simplify the creation of each plugin metric. The output of these functions can be passed as parameters directly to the \codeinline{EvaluationPlugin}.

\paragraph{Evaluation Plugin}
\codeinline{EvaluationPlugin} is the component responsible for the orchestration of both plugin metrics and loggers. Its role is to gather metrics outputs and forward them to the loggers during training and evaluation loops. All the user has to do to keep track of an experiment is to provide the strategy object with an instance of the \codeinline{EvaluationPlugin} with the target metrics and loggers as parameters. Fig.~\ref{code:eval-plugin} shows how to use the evaluation plugin and metric helper functions. \\ 

Avalanche's effort to monitor different facets of performance aims at enabling a wider experimental assessment, which is too often focused only on the forgetting of previous knowledge~\cite{diaz-rodriguez2018}.

\begin{figure}[ht]
\begin{tcolorbox}[title={Evaluation Plugin}, colback=mygray]
\begin{minted}{python}
text_logger = TextLogger(
    open('out.txt', 'w'))
interactive_logger = InteractiveLogger()
tensorboard_logger = TensorboardLogger()

eval_plugin = EvaluationPlugin(
    accuracy_metrics(experience=True),
    loss_metrics(minibatch=True, 
        epoch=True, stream=True),
    ExperienceForgetting(),
    StreamConfusionMatrix(),
    cpu_usage_metrics(stream=True),
    timing_metrics(minibatch=True),
    ram_usage_metrics(epoch=True),
    gpu_usage_metrics(gpu_id, epoch=True),
    disk_usage_metrics(epoch=True),
    MAC_metrics(minibatch=True),
    
    loggers=[interactive_logger, 
             text_logger,
             tensorboard_logger])
\end{minted} 
\end{tcolorbox}
\caption{Avalanche evaluation plugin (or \emph{evaluator}) object instantiation example.}
\label{code:eval-plugin}
\end{figure}

\subsection{Logging} \label{sec:logging}
\begin{figure}[t]
    \centering
    \begin{subfigure}[t]{0.22\textwidth}
        \centering
        \includegraphics[width=\textwidth]{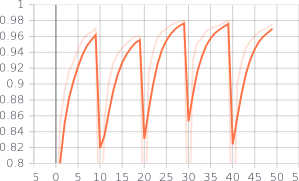}
        \caption{Accuracy over epochs during training}
    \end{subfigure}
    \begin{subfigure}[t]{0.22\textwidth}
        \centering
        \includegraphics[width=\textwidth]{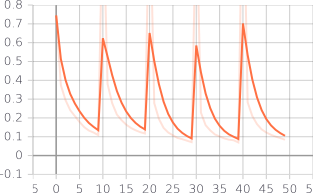}
        \caption{Loss over epochs during training}
    \end{subfigure}
    \begin{subfigure}[t]{0.22\textwidth}
        \centering
        \includegraphics[width=\textwidth]{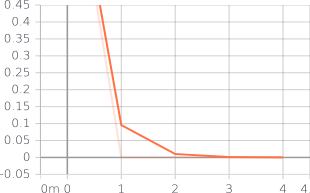}
        \caption{Accuracy over first experience across $5$ training experiences}
    \end{subfigure}
    \begin{subfigure}[t]{0.2\textwidth}
        \centering
        \includegraphics[width=\textwidth]{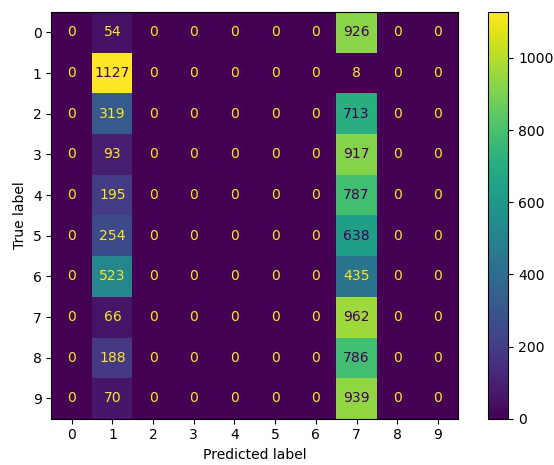}
        \caption{Confusion Matrix}
    \end{subfigure}
    \centering
    \caption{\codeinline{Tensorboard Logger} output examples.}
    \label{fig:tensorboard}
\end{figure}

Nowadays, logging facilities are fundamental to monitor in real time the behavior of an ongoing experiment (which may last from minutes to days). \\
The Avalanche \codeinline{logging} module is in charge of displaying to the user the result of each plugin metric during the different experiment phases. Avalanche provides three different loggers: \codeinline{TextLogger}, \codeinline{InteractiveLogger} and \codeinline{TensorboardLogger} \cite{tensorflow2015-whitepaper}. They provide reports on textual file, standard output and Tensorboard, respectively. As soon as a metric emits a value, the Text Logger prints the complete metric name followed by its value in the destination file. See Fig.~\ref{fig:tensorboard} for an example of Tensorboard output. The \texttt{InteractiveLogger} reports the same output as \texttt{TextLogger}, but it also uses the \emph{tqdm} package\footnote{\url{https://tqdm.github.io}} 
to display a progress bar during training and evaluation. \texttt{TensorboardLogger} is able to show images and more complex outputs, which cannot be appropriately printed on standard output or textual file. We are also working on the integration of the Weights and Biases logger~\cite{wandb}, which should be released soon.\\
Integrating loggers into both training and evaluation loops is straightforward. Once created, loggers have to be passed to the \texttt{EvaluationPlugin}, which will be in charge of redirecting metrics outputs to each logger during the experiment. See Fig.~\ref{code:eval-plugin} for an example of loggers creation.

Users can easily design their own loggers by extending the class \texttt{StrategyLogger}, which provides the necessary interface to interact with the \texttt{EvaluationPlugin}.


\subsection{Models}
\label{sec:models}

The Avalanche \codeinline{models} module offers a set of simple machine learning architectures ready to be used in experiments. In particular, the module contains versions of feedforward and convolutional neural networks and a pre-trained version of MobileNet (v1)~\cite{howard2017}. The main purpose of these models is to let the user focus on Avalanche features, rather than on writing lines of code to build a specific architecture. We plan to extend our model support with more advanced architectures, also tailored to specific CL applications. 

\section{Related Works}
\label{sec:related}

Reproducibility is one of the main principles upon which Avalanche is based. Experiments in the continual learning field are often challenging to reproduce, due to the different implementations of protocols, benchmarks and strategies by different authors. This issue of insufficient reproducibility is not limited to continual learning. The whole artificial intelligence community is affected; a number of authors have recently discussed some possible solutions to the problem~\cite{dToolAI-rep, Pineau2020ImprovingRI, a_step_towards_repro}.

The advent of machine (and deep) learning libraries, mainly TensorFlow \cite{tensorflow2015-whitepaper} and PyTorch \cite{paszke2019pytorch} has partially mitigated the reproducibility problem.
Using these libraries assures a standard implementation of many machine learning building blocks, reducing ambiguities due to bespoke and different implementations of basic concepts. 

In recent times, the continual learning community has put a lot of effort into addressing these problems, by providing code and libraries aimed to increase the reproducibility of continual learning experiments \cite{delange2021continual, cl_benchmarks, masana2020class, pmlr-v80-serra18a-hat, vandeven2018generative, vandeven2018a}. On the one hand, these first attempts lack the generality and the consistency of Avalanche, especially regarding the creation of different and complex benchmarks, and the continual support of a large community. On the other hand, they demonstrate, however, the growing interest of the entire community towards these issues.

Another area other than continual learning which has recently seen a proliferation of libraries and tools similar in spirit to Avalanche is reinforcement learning (RL). One of the most popular such benchmark RL libraries is OpenAI Gym~\cite{openAI-gym}, within which a multitude of different RL environments is available. A similar library is ViZDoom~\cite{wydmuch2018vizdoom}, in which an agent plays the famous computer game \emph{Doom}. Other relevant projects in the field of reinforcement learning are Dopamine~\cite{castro18dopamine}, which focuses on simplicity and easy prototyping, and project Malmo~\cite{malmo}, which is based on the famous Minecraft game. 

Many of these libraries, however, only focus on the agent's interaction with the environment (which, in the continual learning domain, can be translated into the definition of benchmarks and scenarios), providing none or just a few base strategies or baselines. In the reinforcement learning field, this problem is addressed by other libraries that include standard implementations of baselines algorithms, such as OpenAI baselines~\cite{openAI-baselines} and stable baselines \cite{stable-baselines}. 

Another prominent example of a collection of baseline training strategies and pretrained models is the natural language processing transformers library by Hugging Face~\cite{hugginface}. Many basic concepts upon which Avalanche is based (e.g. plugins, loggers, benchmarks) can also be found in more general machine learning libraries such as PyTorch Lighting~\cite{falcon2019pytorch} and fastai~\cite{howard2020fastai}. Indeed, Avalanche is based on the same comprehensiveness and consistency principle, hence not only benchmarks but also strategies and metrics are included. This promotes consistency across the different parts of the library and simplifies the interaction between different modules. Moreover, Avalanche could be integrated with the mentioned deep learning libraries, thanks to the similarity regarding design principles and code structure. 

Another important problem in research is the bookkeeping of experiments. Having a tool that keeps track of any single run (hyper-parameters, model used, algorithm used, variants, inputs to the model, etc.) is crucial for reproducibility, especially when strategies such as grid search are applied.
As discussed in Sec. \ref{sec:logging}, Avalanche already implements a fine-grained and punctual logging, which allows to visualize and save the results of different experiments. Moreover, Avalanche could be easily integrated with standalone libraries specifically developed for experiments bookkeeping and visualization, such as Sacred~\cite{sacred} or Weights and Biases (wandb) \cite{wandb}.

\section{Conclusion and Future Work}
\label{sec:conclusion}

In the last decade, we have witnessed a significant effort towards making research in machine learning more transparent, reproducible and open-access. However, although research papers are increasingly accompanied by publicly hosted codebases, it is often difficult to run and integrate such software into environments which are typically different from the one within which it was originally designed. This not only hampers reproducibility but also inhibits scalability, for each research paper ends up creating its own implementation almost from scratch.

Avalanche aims to provide a coherent, end-to-end, easily extendable library for continual learning research and development; a solid reference point and shared resource for researchers and practitioners working on continual learning and related areas.

As reported in Table~\ref{tab:features}, the current Avalanche \texttt{Alpha} version focuses on continual supervised learning for vision tasks, as a significant amount of deep learning research in this area was designed and assessed in this context \cite{hadsell2020}. However, being Avalanche a \emph{community-driven} effort, we plan in both the near and medium terms to support the integration of additional learning paradigms (e.g. Reinforcement or Unsupervised Learning), tasks type (e.g. Detection, Segmentation) and application contexts (e.g. Natural Language Processing, Speech Recognition), depending on the research community demands and priorities.

We hope that this library, as a powerful avalanche, may trigger a positive reinforcement loop within our community, nudging it to shift towards a more collaborative and inclusive research environment and helping all of us tackle together the grand research challenges presented by a frontier topic such as continual learning.

{\small
\bibliographystyle{ieee_fullname}
\bibliography{egbib, CL}
}

\end{document}